\documentclass[11pt,a4paper]{article}
\usepackage[hyperref]{naaclhlt2019}
\usepackage{times}
\usepackage{latexsym}

\usepackage{url}

\usepackage{url}
\usepackage[ruled,vlined,linesnumbered]{algorithm2e}
\usepackage{amsmath}
\usepackage{amsmath}
\usepackage{amsfonts}
\usepackage{amssymb}
\definecolor{purple}{HTML}{A680B8}
\definecolor{gold}{HTML}{009900}
\definecolor{black}{HTML}{3333FF}
\definecolor{red}{HTML}{FF0000}
\usepackage{graphicx}

\aclfinalcopy 

\title{Medical Entity Linking using Triplet Network}

\author{$^1$Ishani Mondal \qquad $^1$Sukannya Purkayastha \qquad $^1$Sudeshna Sarkar \qquad $^1$Pawan Goyal \\ $^2$\textbf{Jitesh K Pillai \qquad $^2$Amitava Bhattacharyya \qquad $^2$Mahanandeeshwar Gattu} \\ $^1$Department of Computer Science and Engineering, IIT Kharagpur \\ $^2$Excelra Knowledge Solutions Pvt Ltd, Hyderabad, India}

\begin{document}
\maketitle
\begin{abstract}
  Entity linking (or Normalization) is an essential
task in text mining that maps the entity
mentions in the medical text to standard entities in a
given Knowledge Base (KB). This task is of
great importance in the medical domain. It
can also be used for merging different medical and clinical
ontologies. In this paper, we center around the
problem of disease linking or normalization.
This task is executed in two phases: candidate
generation and candidate scoring. In this paper,
we present an approach to rank the candidate
Knowledge Base entries based on their
similarity with disease mention. We make use
of the Triplet Network for candidate ranking.
While the existing methods have used carefully
generated sieves and external resources
for candidate generation, we introduce a robust
and portable candidate generation scheme that
does not make use of the hand-crafted rules.
Experimental results on the standard benchmark
NCBI disease dataset demonstrate that
our system outperforms the prior methods by
a significant margin.
\end{abstract}

\section{Introduction}

A disease is an abnormal medical condition that poses a negative impact on the organisms and enabling access to disease information is the goal of various information extraction as well as text mining tasks. The task of disease normalization consists of assigning a unique concept identifier to the disease names occurring in the clinical text. However, this task is challenging as the diseases mentioned in the text may display morphological or orthographical variations, may utilize different word orderings or equivalent words. Consider the following examples:
\begin{center}
  \fbox{\begin{minipage}{0.45\textwidth}
\textbf{Example 1}: ``..characteristics of the disorder include a \textbf{short trunk and extremities}..."
\textbf{Source} : (PMID:7874117)

\textbf{Example 2}: ``\textbf{Renal amyloidosis}, prevented by colchicine, is the most severe complication of FMF ..."
\textbf{Source} : (PMID:10364520)
\end{minipage}}  
\end{center}
In Example 1, the disease mention \textbf{short trunk and extremities} should be mapped to a candidate Knowledge Base entry(D006130) containing synonyms like \textbf{Growth Disorder}. In Example 2, \textbf{Renal amyloidosis} should be assigned to Knowledge Base ID (C538249) which has synonyms such as, \textbf{Amyloidosis 8}.

Based on our studies and analysis of the medical literature, it has been observed that the same disease name may occur in multiple variant forms such as. synonyms replacement (e.g.\textit{``lung cancer''}, \textit{``lung carcinoma''}), spelling variation (\textit{``Acetolysis''}, \textit{``acetolisis''}), a short description modifier precedes the disease name (e.g. \textit{``massive heart attack"}), different word orderings (eg. \textit{``alpha-galactosidase deficiency"}, \textit{``deficiency of alpha-galactosidase"}). 

In this paper, we have formulated the task of learning mention-candidate pair similarity using Triplet Networks \cite{Hoffer2015DeepML}. Furthermore, we have explored in-domain word\footnote{http://evexdb.org/pmresources/vec-space-models/} and subword embeddings \cite{bojanowski2017enriching} as input representations. We find that sub-word information boosts up the performance due to gained information for out-of-vocabulary terms and word compositionality of the disease mentions.

The primary contributions of this paper are three-fold:  1) By identifying positive and negative candidates concerning a disease mention, we optimize the Triplet Network with a loss function that influences the relative distance constraint  2) We have explored the capability of in-domain sub-word level information\footnote{https://github.com/ncbi-nlp/BioSentVec.git} in solving the task of disease normalization. 3) Unlike existing systems \cite{DSouza+Ng:15b}, \cite{Li2017}, we present a robust and portable candidate generation approach without making use of external resources or hand-engineered sieves to deal with morphological variations. Our system achieves state-of-the-art performance on NCBI disease dataset \cite{Dogan2014}

\section{Dataset}

The NCBI disease corpus \cite{Dogan2014} contains 792 Pubmed abstracts with disorder concepts manually annotated.
In this dataset, disorder mentions in each abstract are manually annotated with the identifier of the concept in the reference ontology to which it refers.
It uses MEDIC lexicon \cite{Davis2012} as the reference ontology. (See Table 1 for dataset statistics)

\begin{table}[t!]
\begin{center}
\begin{tabular}{|l|l|l|l|}
\hline \bf Dataset & \bf Abstracts & \bf Total & \bf  Unique \\ 
\hline
Training set & 692 & 5932 & 1538 \\
Test set & 100 & 960 & 427 \\
Total & 792 & 6892 & 1965 \\
\hline
\end{tabular}
\end{center}
\caption{\label{font-table} NCBI Disease Corpus Statistics }
\end{table}

\section{Methodology}

The dataset consists of a certain number of abbreviations,  in order to identify these cases, we have considered the mentions composed of all upper-case letters as abbreviations. We find the disease mentions immediately preceding the abbreviated terms and substitute all the occurrences of the abbreviated words in that document with the corresponding expanded disease mentions.
Our system primarily consists of two modules: 
1) \textbf{Candidate generation}: (See section 3.1) Generate potential candidates from the Knowledge Base corresponding to a disease mention.  
2) \textbf{Candidate ranking}:  (See section 3.2) Rank those potential candidates corresponding to a disease mention.

\begin{figure*}
\centering
\framebox{
\includegraphics[width=0.8\textwidth,height=6cm]{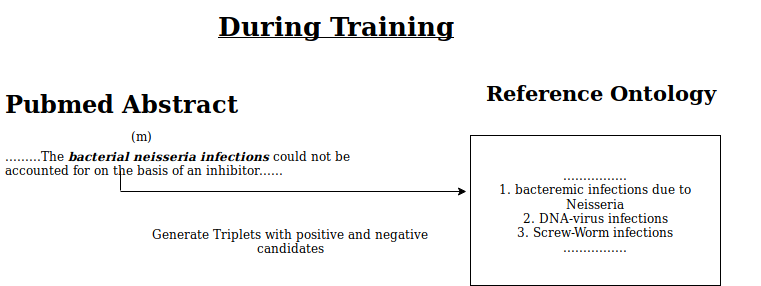}}
\caption{Pictorial Representation of the training Data Generation Process} \label{fig1}
\end{figure*}

\subsection{Candidate generation}
In this section, we discuss the algorithm which generates the potential candidates to which the disease mentions might be referring. In this study, the Knowledge Base entries were sampled from the entire MEDIC Lexicon, but not limited to only annotations in the NCBI Disease Corpus.

For a given disease mention, the candidate generation algorithm generates candidates from the Knowledge Base entries. Suppose, the Knowledge Base consists of $k$ entries, each having a certain number of synonyms. Each multi-word synonym represented by the sum of its word embeddings. For a given  mention $m$ consisting of $l$ words represented by $\{m_{1},m_{2},\dots,m_{l}\}$, we represent $m$ as the sum of its word embeddings. The steps for the candidate generation algorithm are as follows:
\begin{itemize}
  \item \textbf{Step 1}: Candidate Set 1, $\{C_1\}$ : Calculate the cosine similarity between vector representation of each synonym (candidate) of the KBIDs and the mention. Identify the top $k_1$ ids whose candidates have cosine similarity greater than or equal to threshold $t_1$.
  \item \textbf{Step 2}: Candidate Set 2, $\{C_2\} $: Calculate the Jaccard overlap of the mention and the candidates of each KBID. Identify the top $k_2$ ids having Jaccard overlap score greater than or equal to threshold $t_2$.

\end{itemize}

In our experiments, we choose $t_1$=0.7, $t_2$=0.1, $k_1$=3 and $k_2$=7.

We provide examples of candidates generated from our proposed algorithm below.

\begin{center}
  \fbox{\begin{minipage}{0.45\textwidth}
\textbf{Mention}: ``bacteremic infections due to Neisseria"

\textbf{Candidate Set 1, \{$C_1$\}} = \{``bacterial neisseria infections"\}

\textbf{Candidate Set 2, \{$C_2$\}} = \{``bacterial neisseria infections" , ``DNA-virus infections" , ``Screw-Worm Infections" \}
\end{minipage}}  
\end{center}

\subsection{Candidate Ranking}
Assume that there are $n$ candidates represented by $\{c_{1},c_{2},$\dots$,c_{n}\}$ for an entity mention $m$, we use a Triplet Network which has proven to perform well in many Computer Vision \cite{Hoffer2015DeepML} as well as Natural Language Processing tasks \cite{Coref_mention_ranking} . As such given a triplet, the idea is to leverage the notion of reducing the distance between the mention and its positive candidate while increasing the distance between the mention and its negative candidate.

\subsubsection{Triplet data generation}

In order to learn better semantic representations between a disease mention and its corresponding candidates, we have generated training data in the form of triplets consisting of disease mention $m$, positive candidate $q_p$, negative candidate $q_n$. 

An example of the triplet data is given below: 

\begin{center}
  \fbox{\begin{minipage}{0.45\textwidth}
\textbf{Disease Mention}: ``bacteremic infections due to Neisseria"

\textbf{Positive Candidate}: ``bacterial neisseria infections"

\textbf{Negative Candidates}: ``DNA-virus infections", ``Screw-Worm Infections".

The triplets are as follows:

\begin{itemize}
    \item (``bacterial neisseria infections", ``bacteremic infections due to Neisseria", ``DNA-virus infections")  
    \item (``bacterial neisseria infections", ``bacteremic infections due to Neisseria", ``Screw-Worm Infections")  
\end{itemize}
\end{minipage}}  
\end{center}



\begin{figure}[!ht]
\centering
\includegraphics[width=0.5\textwidth,height=6cm]{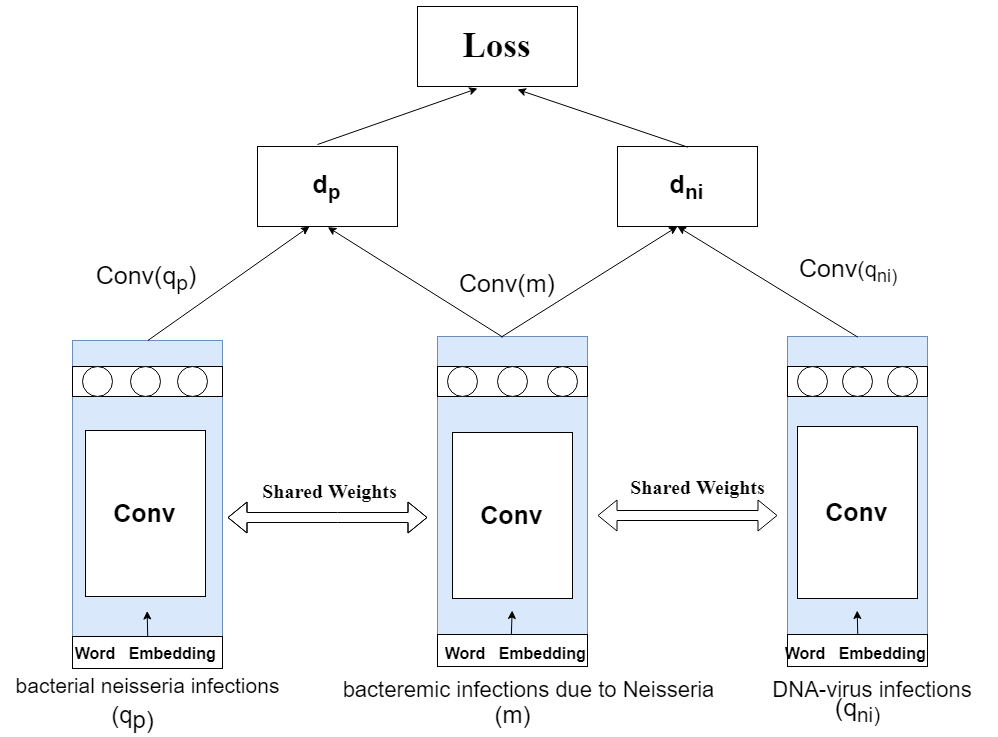}
\caption{The word and sub-word embeddings of triplet (`bacterial neisseria infections', `bacteremic infections due to Neisseria', 'Screw-Worm Infections') are fed as batches into the Triplet Network.} \label{fig2}
\end{figure}

\subsubsection{Model Architecture}
The Triplet Network architecture as proposed by \cite{Hoffer2015DeepML} has been adopted for the task of entity normalization. To train the model, each triplet consisting of mentions and its candidates are fed into the parameter-shared network ($Conv$), as a sequence of word embeddings. For a triplet, ($q_p$, $m$, $q_{n_i}$) the layer outputs their representations $Conv$($q_p$), $Conv$($m$) and $Conv$($q_{n_i}$) respectively. Our objective is to make the representations of $m$ and $q_p$ closer than the representations of $m$ and $q_{n_i}$. Thus the next layer uses a distance function, denoted by $dis$, to compute the distances as follows:
\[ d_p = dis(Conv(m), Conv(q_p)) \]
\[ d_{n_i} = dis(Conv(m), Conv(q_{n_i})) \]

Here $d_p$ specifies the distance between target disease mention $m$ and $q_p$ while $d_{n_i}$ specifies the distance between target disease mention $m$ and $q_{n_i}$.
The triplet loss function ($L$) used for achieving this goal has been formulated as follows:
\[ L = \max{(d_p - d_{n_i} + \alpha, 0)} \]

Another variable $\alpha$, a hyperparameter is added to the loss equation which defines how far away the dissimilarities should be. Thereafter, by using this loss function, we calculate the gradients and update the parameters of the network based on these gradient values. For training the network, we take mention $m$ and randomly sample $q_p$ and $q_{n_i}$ and compute their loss function and update their gradients.

We use 200-dimensional word2vec \cite{Mikolov13} embeddings trained on Wikipedia and Pubmed PMC-Corpus \cite{bioword} as input
to $Conv$. To deal with the huge number of out-of-
vocabulary terms in the medical domain,
we have used the $fastText$ based sub-word embeddings \cite{Galea2018SubwordII}. $fastText$ \cite{bojanowski2017enriching} has been applied on PubMed and MIMIC-III \cite{Johnson2016} to generate 200-
dimensional word embeddings, the window size being 20, learning rate 0.05, sampling threshold
1e-4, and negative examples 10 \cite{zhang2018}.

\subsubsection{Training Details}

$Conv$ is composed of one convolutional and
max-pooling layer. ReLU non-linearity is applied between two consecutive layers. The final embedding of $Conv$ is a fixed-length($128$) vector. For $dis$ and the loss function
we use the L2 distance \cite{DANIELSSON1980227}. The triplet loss has been applied. For training we use Adam Optimizer \cite{Kingma2015AdamAM} with an initial learning rate of 0.001. Training has been done for
50 epochs, and early stopping has been employed
on the basis of the accuracy of the validation set.
After hyperparameter tuning, several experiments
have been performed, and the results on the best
hyperparameter settings have been reported.

\subsubsection{Evaluation}
After the model has been trained, we evaluate the rank of each of the disease mentions in the test set. For each of the disease mention $m$ in the test set, we run the candidate generation algorithm to find out the maximum cosine similar candidates for the potential KnowledgeBaseIDs. The positive candidate is labelled as $1$ while the rest has been labelled as $0$. During the process of evaluation, we calculate the similarity score between the disease mention and its candidates. The similarity scores are then sorted in descending manner in order to rank the candidates based on its similarity. We choose the candidate with the maximum similarity score for each of the disease mentions.

We choose the evaluation measure as accuracy. Since, the highest similar candidate is of primary interest in the task of entity linking, so we choose the top-$K$ ( Where $K=1$). 

$TP$ = It signifies that the highest ranked candidate for disease mention $m$ is the actual referent KnowledgebaseID.

$FP$ = It signifies that the highest ranked candidate for disease mention $m$ is not the actual referent KnowledgebaseID.

$$\textit{Accuracy} = \frac{TP}{TP+FP} $$

\section{Results}
We report accuracy for our system in finding the correct Knowledge Base ID corresponding to a disease mention in the text. \textbf{Table 2} shows that in comparison with the existing baseline systems, \textbf{Subword information} as input to the Triplet Network and abbreviation expansion from the document context (Triplet CNN+subword+abb) performs the best. From the feature ablation, it is clear that the in-domain word embeddings((Triplet CNN + dynamic word2vec) and (Triplet CNN + static word2vec)) are essential for capturing better semantic representations. 

\begin{table}[t!]
\centering
\begin{tabular}{|l|l|}
\hline
  \textbf{Model Name}  & \textbf{Accuracy}\\
  \hline
  \cite{DSouza+Ng:15b} & 84.65 \\
  \cite{Li2017} & 86.10 \\
  \hline
  Triplet CNN + static word2vec & 86.09 \\
  Triplet CNN  + dynamic word2vec & 87.85 \\
  Triplet CNN + subword & 89.65 \\
  Triplet CNN + subword + abb & \textbf{90.01} \\
   \hline
\end{tabular}
\caption{The table shows the accuracy of our system in comparison with the baseline systems.}
\end{table}

\section{Analysis}

In this section, we throw some light on both the merits and demerits of the proposed system with respect to the baseline models.

\subsection{Merit Analysis}


We compare our results with other rule-based and neural network based methods known to perform well on this standard dataset. To gain more insights into our proposed model, in particular, the importance of the domain-specific word and sub-word representation to capture the semantic and syntactic similarity using Triplet Network, we select some examples from the labeled test set.  In figure 2,  two different cases have been shown which demonstrate the performance gap between our and the existing baseline systems.

In Example 1, the disease mention \textit{``inherited neurodegeneration''} was not mapped with \textit {``heredodegenerative disorders''} ( D020271) by the existing methods, because of their incapability to capture the semantic similarity. In contrast to this, our system obtains additional semantic and syntactic information from the domain-specific sub-word embeddings and thereby maps to the correct concept ID. 

In Example 2, the abbreviation \textit{``AS''} is polysemous in nature as it can either be mapped to the concepts like \textit{``Angelman Syndrome''} (PMID : 9585605 ) and \textit{``Ankolysing Spondylitis''} ( PMID : 9336417). Due to the lack of contextual information in the existing models, they were not able to handle the polysemous nature of the abbreviations; but abbreviation expansion from the document level context in our system handles this scenario.

\begin{figure}[!h]
  \includegraphics[width=0.45\textwidth,height=5cm]{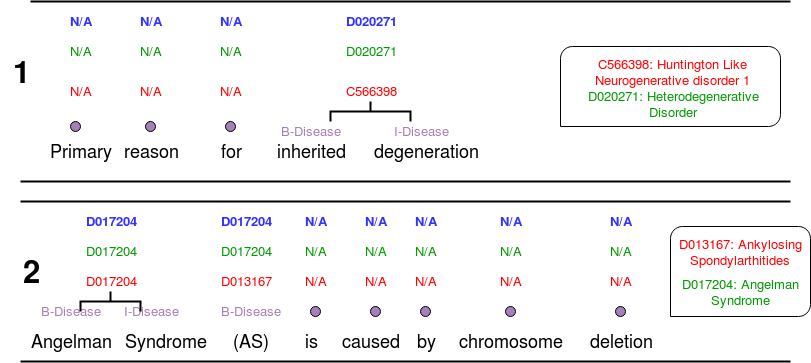}
  \caption{The NER tags as input are shown in \textcolor{purple}{purple}, the gold standard conceptID is shown in \textcolor{gold}{green}, the predictions from the baseline systems are shown in \textcolor{red}{red}, whereas the prediction from the proposed system is shown in \textcolor{black}{blue}.}
\end{figure}

\subsection{Demerit Analysis}

The error types incurred by the proposed system have been explained in detail as follows:

1) \textbf{Ambiguous distribution of importance to the disease name}: The system fails to understand which part of disease mention to provide more attention while performing normalization. Suppose the disease mention is \textbf{``colorectal adenoma"}, during normalization, the system mistakenly normalizes the disease to the concept ID predominated by \textbf{``colorectal''}. Automatic identification of such semantic attention is challenging and deserves a significant spot in the future research.

2) \textbf{Incorrect mapping of certain ambiguous disease names}: Suppose the disease mention dysmorphic features in ``..loss of MAGEL2 may be critical to abnormalities in brain development and \textbf{dysmorphic features} in individuals with PWS.." ( PMID: 10915770 ) has been mapped to D057215 whereas the same disease mention in ``..She had minor \textbf{dysmorphic features}
consistent with those of.." ( PMID: 8071957 ) has been assigned to D000013. Since, in these two examples, the disease mention in these two examples have been assigned as "diseaseClass" and "Modifier" features respectively. It happens due to different NER features of the mention annotated in the dataset. But incorporating this NER feature in our proposed model unnecessarily generates huge number of false positives.

\section{Conclusion}

In this paper, we have formulated the task of entity linking as a candidate ranking approach. Using a Triplet Network,  we learn high-quality representations of candidates, tailored to reveal relative distances between the disease mention and its positive and negative candidates. Furthermore,  we take a step towards eliminating the need to generate candidates based on hand-crafted rules and external knowledge resources. Though our method outperforms the existing systems by a strong margin, there is a scope for improvement in terms of attention-based disease similarity  (viz, ``Neisseric infections'' imply the importance of ``Neisseric'' during its similarity computation with the ``bacterial neisseria infections''). An intriguing course for future work is to further explore the robustness and scalability of this approach to other clinical datasets for entity normalization.

\section*{Acknowledgments}

This work has been supported by the project ``Effective Drug Repurposing through literature and patent mining, data integration and development of systems pharmacology platform'' sponsored by MHRD, India and Excelra Knowledge Solutions, Hyderabad. Besides, the authors would like to thank the anonymous reviewers for their valuable comments and feedback.

\bibliography{naaclhlt2019}
\bibliographystyle{naaclhlt2019}

\end{document}